\documentclass[runningheads]{llncs}
\usepackage{graphicx}
\usepackage{amsmath}
\usepackage{booktabs}
\usepackage{array}
\newcolumntype{H}{>{\setbox0=\hbox\bgroup}c<{\egroup}@{}}

\begin{document}
\title{Billions of Parameters Are Worth More Than In-domain Training Data:\\A case study in the Legal Case Entailment Task}
%
%
\author{Guilherme Moraes Rosa \and
Luiz Bonifacio \and
Vitor Jeronymo \and \\
Roberto Lotufo \and Rodrigo Nogueira}
\authorrunning{Rosa et al.}
\titlerunning{Billions of Parameters Are Worth More Than In-domain Training Data}
%
\institute{NeuralMind, Brazil \\
University of Campinas (UNICAMP), Brazil}
\maketitle              
\begin{abstract}

Recent work has shown that language models scaled to billions of parameters, such as GPT-3, perform remarkably well in zero-shot and few-shot scenarios. In this work, we experiment with zero-shot models in the legal case entailment task of the COLIEE 2022 competition. 
Our experiments show that scaling the number of parameters in a language model improves the F1 score of our previous zero-shot result by more than 6 points, suggesting that stronger zero-shot capability may be a characteristic of larger models, at least for this task. Our 3B-parameter zero-shot model outperforms all models, including ensembles, in the COLIEE 2021 test set and also achieves the best performance of a single model in the COLIEE 2022 competition, second only to the ensemble composed of the 3B model itself and a smaller version of the same model. Despite the challenges posed by large language models, mainly due to latency constraints in real-time applications, we provide a demonstration of our zero-shot monoT5-3b model being used in production as a search engine, including for legal documents. The code for our submission and the demo of our system are available at~\url{https://github.com/neuralmind-ai/coliee} and ~\url{https://neuralsearchx.neuralmind.ai}, respectively.


\end{abstract}
%
%
\section{Introduction}

A common practice in machine learning competitions and leaderboards is to use training and test data from the same distribution. However, winning models often fail when applied to real-world applications, since real-world data is usually more diverse and may differ from the data seen during training \cite{domain1,domain2}. Thus, our goal is to develop models capable of generalizing to examples outside of the training data distribution. 

In the legal domain, it is known that the low availability of annotated data is a serious problem \cite{CUAD,Ratnayaka_2020}, making it very difficult to develop high-performance models for legal tasks. An alternative to address this problem is to transfer knowledge using models fine-tuned on related domains where annotated data is abundant. However, models tend to underperform in scenarios that involve different data distributions (e.g., text domains) between training and inference \cite{domain_adaptation,domain_adaptation2}. Developing models capable of overcoming this instability is an important challenge to be addressed, as in the long term it is desirable that our models are robust enough to take advantage of previously learned information to generalize well to new domains and tasks.

It has also been shown that the high number of parameters in larger language models, up to billions, results in greater generalization ability and consequently superior zero-shot performance compared to smaller models~\cite{NEURIPS2020_1457c0d6,wei2022finetuned,sahn_2021}. Rabelo et al.~\cite{rabelo2020} were the first to apply zero-shot techniques to the legal case entailment task, and Rosa et al.~\cite{icail_2021} showed that zero-shot models can outperform fine-tuned ones on this task. In this work, we show that performance on the legal case entailment task can be further improved simply by scaling the language model, without the need for any adaptation to the legal domain or fine-tuning on the in-domain training dataset. 

\section{Related Work}

Zero-shot and few-shot models have demonstrated competitive performance compared to fine-tuned ones in many textual tasks. For example, Pradeep et al. \cite{pradeep2020h2oloo} used a Text-To-Text Transfer Transformer (T5) model fine-tuned only on a general domain ranking dataset to achieve the best or second-best performance in multiple domain-specific tasks such as Precision Medicine~\cite{Roberts2019OverviewOT} and TREC-COVID~\cite{zhang2020rapidly}. Rosa et al. \cite{icail_2021} showed for the legal case entailment task that language models without any fine-tuning on the target task can perform better than models fine-tuned on the task itself. They argued that, given limited labeled data for training and a held-out test set, such as in the COLIEE competition, a zero-shot model fine-tuned only on a large and diverse general domain dataset may be able to generalize better to unseen data than models fine-tuned on a small domain-specific dataset. In this work, we evaluate this zero-shot approach, but using a much larger model.

Larger language models have recently shown stronger zero-shot and few-shot capabilities than smaller models. For example, Brown et al. \cite{NEURIPS2020_1457c0d6} presented GPT-3, a language model with 175 billion parameters. They evaluated the model learning ability in few-shot scenarios and showed that scaling up language models greatly improves performance, sometimes even achieving competitive results compared to models fine-tuned on hundreds of thousands of examples. Sanh et al. \cite{sahn_2021} developed a system to map any natural language task into human-readable prompted forms and converted several supervised datasets into that format. They fine-tuned a T5 model of 11 billion parameters on this new multitask dataset and found that the model achieves strong zero-shot performance across multiple held-out datasets, including outperforming models up to 16 times its size. Wei et al. \cite{wei2022finetuned} introduced \textit{instruction tuning}, a new approach to improve the zero-shot learning ability of larger language models, in which task-specific instructions are given to models using natural language. This approach substantially improved performance and outperformed the GPT-3 model on multiple datasets in zero-shot scenarios.

Zhong et al. \cite{metalmm} proposed \textit{meta-tuning}, a method for optimizing zero-shot learning of large language models by fine-tuning on a collection of datasets in a question answering format. The model outperformed question answering models of the same size, including the previous state-of-the-art method. Furthermore, recent results suggest that we still did not reach the limits of this trend of increasing model size. To understand the impact of scaling in few-shot learning, Chowdhery et al. \cite{PALM} introduced the Pathways Language Model (PaLM), a language model with 540 billion parameters that achieved state-of-the-art few-shot results in hundreds of language tasks. Furthermore, PaLM 540B outperformed fine-tuned state-of-the-art models in multiple tasks and also surpassed the average human performance on the BIG-bench benchmark.

Winata et al. \cite{winata_2021} evaluated the few-shot learning ability of several GPT and T5 models in multilingual scenarios. They showed that given a few examples in English as context, language models can correctly predict test samples in languages other than English. Lin et al. \cite{lin_2021} fine-tuned a multilingual language model with 7.5 billion parameters in a wide range of languages to study the model's zero-shot and few-shot learning capabilities across multiple tasks. The model achieved state-of-the-art results in more than 20 languages, outperforming GPT-3 on a number of tasks such as multilingual common sense reasoning and natural language inference.

However, these promising results were not limited to models with billions of parameters. Mishra et al. \cite{mishra_2021} fine-tuned a BART model with 140 million parameters using samples containing instructions written in natural language and evaluated the model few-shot ability on unseen tasks. The results suggested that fine-tuning on a collection of tasks improves performance on unseen tasks, even for smaller models.

\section{Legal Case Entailment Task} 

The Competition on Legal Information Extraction/Entailment (COLIEE)~\cite{coliee_overview_2021,rabelo2020coliee,rabelo2019coliee,kano2018coliee,Kano2017OverviewOC} is an annual competition whose goal is to improve and evaluate the performance of automated systems in legal tasks.

Legal case entailment is the second task of COLIEE 2022, which consists of identifying a set of candidate paragraphs from previous judicial decisions that corroborate with or are relevant to a given new judicial decision. 
The dataset comprises case laws collected from the Federal Court of Canada. The training data contains 525 judicial decisions, their respective candidate paragraphs that perhaps may be relevant to the given decision, and a set of labels containing the candidate paragraphs that entail the judicial decision. The test set contains 100 judicial decisions and only their candidate paragraphs, since labels are only revealed after the competition. The dataset has an average of 35 candidate paragraphs per judicial decision, of which only one is relevant on average. In Table~\ref{table:task2}, we show the statistics of the COLIEE 2021 and COLIEE 2022 datasets.

\begin{table}
\caption{COLIEE dataset statistics.}\label{table:task2}
\centering\centering\resizebox{0.85\textwidth}{!}{
\begin{tabular}{l | c | c | c | c } 
 \toprule
 & \multicolumn{2}{c|}{\textbf{2021}} & \multicolumn{2}{c}{\textbf{2022}} \\ 
 & \textbf{ Train } &  \textbf{ Test } & \textbf{ Train } &  \textbf{ Test }  \\
 \midrule
 Examples (base cases) & 425 & 100 & 525 & 100  \\ 
 
 Avg. \# of candidates / example & 35.80 & 35.24 & 35.69 & 32.78 \\
 
 Avg. positive candidates / example & 1.17 & 1.17 & 1.14 & - \\
 
 Avg. of tokens in base cases & 37.51 & 32.97 & 36.64 & 32.21 \\
 Avg. of tokens in candidates & 103.14 & 100.83 & 102.71 & 104.61 \\
 \bottomrule
\end{tabular}
}
\end{table}

The micro F1 score is the official metric in this task:

\begin{equation}
    \mathbf{F1} = \frac{2 \times Precision \times Recall}{Precision + Recall},
\end{equation} \\
\noindent where $Precision$ is the number of candidate paragraphs correctly retrieved for all judicial decisions divided by the number of paragraphs retrieved for all judicial decisions, and $Recall$ is the number of candidate paragraphs correctly retrieved for all judicial decisions divided by the number of relevant candidate paragraphs for all judicial decisions.

\section{Method}  

In this section, we describe monoT5, which is the model used throughout this work.
MonoT5 is an adaptation of the T5 model \cite{raffel2020t5} designed by Nogueira et al. \cite{nogueira2020document}. The authors proposed to fine-tune the model on the MS MARCO dataset \cite{MS_MARCO_v3} to generate the tokens ``true'' or ``false'' depending on the relevance of a document to a query. During inference, monoT5 generates a score that measures this relevance by applying a softmax function to the logits of the tokens ``true'' and ``false'',  and then considering the probability of the token ``true'' as the final score. The model is close to or at the state of the art on datasets such as TREC-COVID \cite{trec-covid}, TREC 2020 Precision Medicine \cite{pradeep2020h2oloo} and Robust04 \cite{trec2004}.

We extend our zero-shot approach developed for the COLIEE 2021 competition \cite{icail_2021}, but now we experiment with a much larger model, which has billions of parameters. We refer to this approach as zero-shot because the models were only fine-tuned on general domain text coming from MS MARCO and directly evaluated on COLIEE's test set, which is made up of legal texts. We use the T5-base and T5-3B models available at Hugging Face,\footnote{\url{https://huggingface.co/castorini/monot5-base-msmarco-10k}, \url{https://huggingface.co/castorini/monot5-3B-msmarco-10k}} and already fine-tuned for 10k steps on MS MARCO, which corresponds to almost one epoch of its training set.
We refer to the resulting models as monoT5-base-zero-shot and monoT5-3b-zero-shot.

At inference time, monoT5 uses the following input template:

\begin{equation}
\text{Query: } \hspace{0.1cm} q \ \ \text{ Document: }  \hspace{0.1cm} d \ \ \text{ Relevant:}
\end{equation}

\noindent where $q$ represents a fragment of a judicial decision from a new legal case and $d$ represents one of the candidate paragraphs of existing legal cases that may or may not be relevant to the given decision.

The model receives the input prompt and estimates a score $s$ that quantifies the relevance of a candidate paragraph $d$ to a fragment of a judicial decision $q$. That is,

\begin{equation}
s = P(\textrm{Relevant}=1 | d, q).
\end{equation}

\noindent The final score for each input is the probability assigned by the model to the token ``true''. After computing all scores, we apply the method described in Section~\ref{section:answer_selection} to select the relevant candidate paragraphs.

\subsection{Answer Selection}   
\label{section:answer_selection}

The models used in the experiments estimate a score for each pair of judicial decision and candidate paragraph.
To select the final set of paragraphs for a given judicial decision, we apply three rules:

\begin{itemize}
    \item Select paragraphs whose scores are above a threshold $\alpha$;
    \item Select the top $\beta$ paragraphs with respect to their scores;
    \item Select paragraphs whose scores are at least $\gamma$ of the top score.
\end{itemize} 

 \noindent We use grid search to find the best values for $\alpha$, $\beta$, $\gamma$ on the training set of the COLIEE 2022 dataset. 
We sweep $\alpha = [0, 0.1, ..., 0.9]$, $\beta = [1, 2 ..., 10]$, and $\gamma = [0, 0.1, ..., 0.9, 0.95, 0.99, 0.995, ..., 0.9999]$.

Note that our hyperparameter search includes the possibility of not using the first rule or the third rule if $\alpha=0$ or $\gamma=0$ are chosen, respectively.

\subsection{Ensemble}

We use the following approach to combine the answers from two models. First, we concatenate the answers provided by each model. Second, we remove duplicates while preserving the highest score. We then apply the answer selection method explained in Section~\ref{section:answer_selection} to select the final set of answers.
The goal is to keep only answers with a certain degree of confidence (i.e., score).

\section{Results}  

We present our results in Table~\ref{table:main_results}.
The BM25 method scores above the median of submissions in the COLIEE 2020 and 2021 datasets (row 2 vs. 1a), suggesting that BM25 is a strong baseline and agrees with the results from other competitions \cite{pradeep2020h2oloo,coliee_retrieval}. The median F1 score for 2022 submissions is considerably higher than in previous years, approaching our best method of 2021, indicating that other teams are using pretrained transformer models for this task.

\begin{table*}[ht]
\caption{\label{tab:task_2} F1 scores on the test sets of the legal case entailment task of COLIEE 2020, 2021 and 2022. Our best single model for each year is in bold.}
\centering\centering\resizebox{1.0\textwidth}{!}{
 \begin{tabular}{l | c | c | c | c | c H } 
 \toprule
\textbf{Description} & \textbf{ Submission name }  & \textbf{ Params } & \textbf{ 2020 } & \textbf{ 2021 } & \textbf{ 2022 }  & $\alpha, \beta, \gamma$  \\
 \hline
 (1a) Median of submissions &  & - & 0.5718 & 0.5860 & 0.6391 & -  \\
 (1b) Best of 2020~\cite{nguyen2020jnlp}  & JNLP.task2.BMWT & - & 0.6753  & - & - & -\\
 (1c) 2nd best team of 2021~\cite{rabelo2021_task2} & UA\_reg\_pp & - & -  & 0.6274 & - & -  \\
 (1d) 2nd best team of 2022 & run2\_bert\_amr & - & -  & - & 0.6694 & -  \vspace{0.1cm} \\
 \midrule  
 (2) BM25 ~\cite{icail_2021} &  & - & 0.6046 & 0.6009 & - &  0.07, 2, 0.99 \\
 (3) DeBERTa ~\cite{icail_2021} & DeBERTa & 350M & \textbf{0.7094} & 0.6339 & - & 0, 2, 0.999 \\ 
 (4) monoT5-base-zero-shot & monot5-base & 220M & - &  0.7192 & 0.6325 & 0, 4, 0.995 \\
 (5) monoT5-large ~\cite{icail_2021} & monoT5 & 770M & 0.6887 &  0.6610 & - & 0, 3, 0.995 \\
 (6) monoT5-large-zero-shot ~\cite{icail_2021} &   & 770M & 0.6577 & 0.6872 & - &  0, 3, 0.995
 \\
 (7) monoT5-3b-zero-shot & monot5-3b & 3,000M & - & \textbf{0.7477} & \textbf{0.6757} &  20, 3, 0.999  \vspace{0.1cm} \\
 \midrule  
 (8) Ensemble of (3) and (5) ~\cite{icail_2021} & DebertaT5 & 1,120M & 0.7217 & 0.6912 & - & 0.6, 2, 0.999 \\
 (9) Ensemble of (4) and (7) & monot5-ensemble & 3,220M & - & 0.7567 & 0.6783 & 0, 3, 0.995  \vspace{0.1cm} \\
 \bottomrule
\end{tabular}
} %

\label{table:main_results}
\end{table*}

Results in the COLIEE 2021 test set show that scaling the model size from base to large leads to a decrease in performance. It is not clear why this happens, but it is not the first time that T5-base achieves better performance than T5-large on an entailment task \cite{carmo2020ptt5}. However, scaling to monoT5-3B provides a significant performance gain, easily outperforming all single models and confirming the trend that much larger models bring better model quality.

For the COLIEE 2022 competition, we submitted three runs using different sizes of monoT5-zero-shot models: a monoT5-base (row 4), a monoT5-3B (row 7) and an ensemble between monoT5-base and monoT5-3B (row 9). Performance consistently increases with model size and two of our submissions (rows 7 and 9) score above the median of submissions and above the second best team in the competition. 
Furthermore, our ensemble method effectively combines the predictions of different monoT5 models, achieving the best performance among all submissions (row 9).

\subsection{Ablation of the Answer Selection Method}

\begin{table*}
\caption{Ablation on the 2021 test set of the answer selection method.}
\centering\centering\resizebox{1.0\textwidth}{!}{
 \begin{tabular}{l | c | c | c | c } 
 \toprule
\textbf{Model} & \textbf{ F1 Score } &  \textbf{ Precision } &  \textbf{ Recall } & $\alpha, \beta, \gamma$  \\
 \midrule  
 monoT5-base-zero-shot (no rule)  & 0.7004  & 0.7600 & 0.6495 & 0, 1, 0\\
 monoT5-base-zero-shot & 0.7192 & 0.7387 & 0.7008 & 0, 4, 0.995  \\
 \midrule
 monoT5-3b-zero-shot (no rule) & 0.7373 & 0.8000 & 0.6838 & 0, 1, 0 \\
 monoT5-3b-zero-shot & 0.7477 & 0.7904 & 0.7094 & 20, 3, 0.999 \\
 \bottomrule
\end{tabular}
}
\vspace{0.1cm}

\label{table:ablation}
\end{table*}

In Table \ref{table:ablation}, we show the ablation result of the answer selection method proposed in Section~\ref{section:answer_selection}.
Our baseline answer selection method uses only the candidate paragraph with the highest score as the final answer set, i.e.,  $\alpha=\gamma=0$ and $\beta=1$.
For all models, the proposed answer selection method gives improvements of at least one F1 point over the baseline.

\section{Conclusion}

In this work, we explored the zero-shot ability of a multi-billion parameter language model in the legal domain. We showed that, for the legal case entailment task, language models without any fine-tuning on the target dataset and target domain can outperform models fine-tuned on the task itself. Furthermore, our results support the hypothesis that scaling language models to billions of parameters improves zero-shot performance. This method has the potential to be extended to other legal tasks, such as legal information retrieval and legal question answering, especially in limited annotated data scenarios.

However, while large language models have achieved state-of-the-art performance in many tasks, one of their main disadvantages is inference speed, primarily due to their computationally intensive nature, which can make them expensive to deploy in real-world applications. Addressing this problem is an important challenge that requires the development of techniques that seek to optimize the latency of the models.

\bibliography{acmart}

\begin{thebibliography}{10}
\providecommand{\url}[1]{\texttt{#1}}
\providecommand{\urlprefix}{URL }
\providecommand{\doi}[1]{https://doi.org/#1}

\bibitem{MS_MARCO_v3}
Bajaj, P., Campos, D., Craswell, N., Deng, L., Gao, J., Liu, X., Majumder, R.,
  McNamara, A., Mitra, B., Nguyen, T., Rosenberg, M., Song, X., Stoica, A.,
  Tiwary, S., Wang, T.: {MS} {MARCO}: {A Human Generated MAchine Reading
  COmprehension Dataset}. arXiv:1611.09268v3  (2018)

\bibitem{domain_adaptation2}
Ben-David, E., Oved, N., Reichart, R.: Pada: Example-based prompt learning for
  on-the-fly adaptation to unseen domains. arXiv preprint arXiv:2102.12206
  (2021)

\bibitem{NEURIPS2020_1457c0d6}
Brown, T., Mann, B., Ryder, N., Subbiah, M., Kaplan, J.D., Dhariwal, P.,
  Neelakantan, A., Shyam, P., Sastry, G., Askell, A., Agarwal, S.,
  Herbert-Voss, A., Krueger, G., Henighan, T., Child, R., Ramesh, A., Ziegler,
  D., Wu, J., Winter, C., Hesse, C., Chen, M., Sigler, E., Litwin, M., Gray,
  S., Chess, B., Clark, J., Berner, C., McCandlish, S., Radford, A., Sutskever,
  I., Amodei, D.: Language models are few-shot learners. In: Larochelle, H.,
  Ranzato, M., Hadsell, R., Balcan, M.F., Lin, H. (eds.) Advances in Neural
  Information Processing Systems. vol.~33, pp. 1877--1901. Curran Associates,
  Inc. (2020),
  \url{https://proceedings.neurips.cc/paper/2020/file/1457c0d6bfcb4967418bfb8ac142f64a-Paper.pdf}

\bibitem{carmo2020ptt5}
Carmo, D., Piau, M., Campiotti, I., Nogueira, R., Lotufo, R.: Ptt5: Pretraining
  and validating the t5 model on brazilian portuguese data (2020)

\bibitem{PALM}
Chowdhery, A., Narang, S., Devlin, J., Bosma, M., Mishra, G., Roberts, A.,
  Barham, P., Chung, H.W., Sutton, C., Gehrmann, S., Schuh, P., Shi, K.,
  Tsvyashchenko, S., Maynez, J., Rao, A., Barnes, P., Tay, Y., Shazeer, N.,
  Prabhakaran, V., Reif, E., Du, N., Hutchinson, B., Pope, R., Bradbury, J.,
  Austin, J., Isard, M., Gur-Ari, G., Yin, P., Duke, T., Levskaya, A.,
  Ghemawat, S., Dev, S., Michalewski, H., Garcia, X., Misra, V., Robinson, K.,
  Fedus, L., Zhou, D., Ippolito, D., Luan, D., Lim, H., Zoph, B., Spiridonov,
  A., Sepassi, R., Dohan, D., Agrawal, S., Omernick, M., Dai, A.M., Pillai,
  T.S., Pellat, M., Lewkowycz, A., Moreira, E., Child, R., Polozov, O., Lee,
  K., Zhou, Z., Wang, X., Saeta, B., Diaz, M., Firat, O., Catasta, M., Wei, J.,
  Meier-Hellstern, K., Eck, D., Dean, J., Petrov, S., Fiedel, N.: Palm: Scaling
  language modeling with pathways. arXiv preprint arXiv:2204.02311  (2022)

\bibitem{domain1}
Dash, T., Chitlangia, S., Ahuja, A., Srinivasan, A.: A review of some
  techniques for inclusion of domain-knowledge into deep neural networks. arXiv
  preprint arXiv:2107.10295  (2021)

\bibitem{domain2}
Farahani, A., Voghoei, S., Rasheed, K., Arabnia, H.R.: A brief review of domain
  adaptation. arXiv preprint arXiv:2010.03978  (2020)

\bibitem{CUAD}
Hendrycks, D., Burns, C., Chen, A., Ball, S.: Cuad: An expert-annotated nlp
  dataset for legal contract review. arXiv preprint arXiv:2103.06268  (2021)

\bibitem{Kano2017OverviewOC}
Kano, Y., Kim, M., Goebel, R., Satoh, K.: Overview of {COLIEE} 2017. In: COLIEE
  2017 (EPiC Series in Computing, vol. 47). pp.~1--8 (2017)

\bibitem{kano2018coliee}
Kano, Y., Kim, M.Y., Yoshioka, M., Lu, Y., Rabelo, J., Kiyota, N., Goebel, R.,
  Satoh, K.: {COLIEE-2018}: Evaluation of the competition on legal information
  extraction and entailment. In: JSAI International Symposium on Artificial
  Intelligence. pp. 177--192 (2018)

\bibitem{rabelo2021_task2}
Kim, M., Rabelo, J., Goebel, R.: Bm25 and transformer-based legal information
  extraction and entailment. Proceedings of the COLIEE Workshop in ICAIL
  (2021)

\bibitem{lin_2021}
Lin, X.V., Mihaylov, T., Artetxe, M., Wang, T., Chen, S., Simig, D., Ott, M.,
  Goyal, N., Bhosale, S., Du, J., Pasunuru, R., Shleifer, S., Koura, P.S.,
  Chaudhary, V., O'Horo, B., Wang, J., Zettlemoyer, L., Kozareva, Z., Diab, M.,
  Stoyanov, V., Li, X.: Few-shot learning with multilingual language models.
  arXiv preprint arXiv:2112.10668  (2021)

\bibitem{mishra_2021}
Mishra, S., Khashabi, D., Baral, C., Hajishirzi, H.: Natural-instructions:
  Benchmarking generalization to new tasks from natural language instructions.
  arXiv preprint arXiv:1611.09268  (2021)

\bibitem{nguyen2020jnlp}
Nguyen, H.T., Vuong, H.Y.T., Nguyen, P.M., Dang, B.T., Bui, Q.M., Vu, S.T.,
  Nguyen, C.M., Tran, V., Satoh, K., Nguyen, M.L.: Jnlp team: Deep learning for
  legal processing in coliee 2020. arXiv preprint arXiv:2011.08071  (2020)

\bibitem{nogueira2020document}
Nogueira, R., Jiang, Z., Pradeep, R., Lin, J.: Document ranking with a
  pretrained sequence-to-sequence model. In: Proceedings of the 2020 Conference
  on Empirical Methods in Natural Language Processing: Findings. pp. 708--718
  (2020)

\bibitem{pradeep2020h2oloo}
Pradeep, R., Ma, X., Zhang, X., Cui, H., Xu, R., Nogueira, R., Lin, J.: H2oloo
  at trec 2020: When all you got is a hammer... deep learning, health
  misinformation, and precision medicine. Corpus  \textbf{5}(d3), ~d2 (2020)

\bibitem{rabelo2020}
Rabelo, J., Kim, M., Goebel, R.: Application of text entailment techniques in
  coliee 2020. International Workshop on Juris-informatics (JURISIN) associated
  with JSAI International Symposia on AI (JSAI-isAI)  (2020)

\bibitem{coliee_overview_2021}
Rabelo, J., Goebel, R., Kim, M.Y., Kano, Y., Yoshioka, M., Satoh, K.: Overview
  and discussion of the competition on legal information extraction/entailment
  (coliee) 2021. Rev Socionetwork Strat (2022).
  https://doi.org/10.1007/s12626-022-00105-z  (2022)

\bibitem{rabelo2019coliee}
Rabelo, J., Kim, M.Y., Goebel, R., Yoshioka, M., Kano, Y., Satoh, K.: A summary
  of the {COLIEE} 2019 competition. In: JSAI International Symposium on
  Artificial Intelligence. pp. 34--49 (2019)

\bibitem{rabelo2020coliee}
Rabelo, J., Kim, M.Y., Goebel, R., Yoshioka, M., Kano, Y., Satoh, K.: Coliee
  2020: methods for legal document retrieval and entailment. In: JSAI
  International Symposium on Artificial Intelligence. pp. 196--210. Springer
  (2020)

\bibitem{raffel2020t5}
Raffel, C., Shazeer, N., Roberts, A., Lee, K., Narang, S., Matena, M., Zhou,
  Y., Li, W., Liu, P.J.: Exploring the limits of transfer learning with a
  unified text-to-text transformer. Journal of Machine Learning Research
  \textbf{21}(140),  1--67 (2020), \url{http://jmlr.org/papers/v21/20-074.html}

\bibitem{domain_adaptation}
Ramponi, A., Plank, B.: Neural unsupervised domain adaptation in nlp---a
  survey. arXiv preprint arXiv:2006.00632  (2020)

\bibitem{Ratnayaka_2020}
Ratnayaka, G., de~Silva, N., Perera, A.S., Pathirana, R.: Effective approach to
  develop a sentiment annotator for legal domain in a low resource setting. In
  Proceedings of the 34th Pacific Asia Conference on Language, Information and
  Computation, pages 252–260, Hanoi, Vietnam. Association for Computational
  Linguistics.  (2020)

\bibitem{Roberts2019OverviewOT}
Roberts, K., Demner-Fushman, D., Voorhees, E., Hersh, W., Bedrick, S., Lazar,
  A.J., Pant, S.: Overview of the trec 2019 precision medicine track. The ...
  text REtrieval conference : TREC. Text REtrieval Conference  \textbf{26}
  (2019)

\bibitem{coliee_retrieval}
Rosa, G.M., Rodrigues, R.C., , Nogueira, R., Lotufo, R.: Yes, bm25 is a strong
  baseline for legal case retrieval. Proceedings of the Eighth International
  Competition on Legal Information Extraction/Entailment  (2021)

\bibitem{icail_2021}
Rosa, G.M., Rodrigues, R.C., Lotufo, R., Nogueira, R.: To tune or not to tune?
  zero-shot models for legal case entailment. ICAIL’21, Eighteenth
  International Conference on Artificial Intelligence and Law, June 21–25,
  2021, São Paulo, Brazil  (2021)

\bibitem{sahn_2021}
Sanh, V., Webson, A., Raffel, C., Bach, S.H., Sutawika, L., Alyafeai, Z.,
  Chaffin, A., Stiegler, A., Scao, T.L., Raja, A., Dey, M., Bari, M.S., Xu, C.,
  Thakker, U., Sharma, S.S., Szczechla, E., Kim, T., Chhablani, G., Nayak, N.,
  Datta, D., Chang, J., Jiang, M.T.J., Wang, H., Manica, M., Shen, S., Yong,
  Z.X., Pandey, H., Bawden, R., Wang, T., Neeraj, T., Rozen, J., Sharma, A.,
  Santilli, A., Fevry, T., Fries, J.A., Teehan, R., Bers, T., Biderman, S.,
  Gao, L., Wolf, T., Rush, A.M.: Multitask prompted training enables zero-shot
  task generalization. arXiv preprint arXiv:2110.08207  (2021)

\bibitem{trec-covid}
Voorhees, E., Alam, T., Bedrick, S., Demner-Fushman, D., Hersh, W.R., Lo, K.,
  Roberts, K., Soboroff, I., Wang, L.L.: Trec-covid: Constructing a pandemic
  information retrieval test collection. arXiv preprint arXiv:2005.04474
  (2020)

\bibitem{trec2004}
Voorhees, E.M.: Overview of the trec 2004 robust track. Proceedings of the
  Thirteenth Text REtrieval Conference, TREC 2004, Gaithersburg, Maryland,
  November 16-19, 2004  (2004)

\bibitem{wei2022finetuned}
Wei, J., Bosma, M., Zhao, V., Guu, K., Yu, A.W., Lester, B., Du, N., Dai, A.M.,
  Le, Q.V.: Finetuned language models are zero-shot learners. In: International
  Conference on Learning Representations (2022),
  \url{https://openreview.net/forum?id=gEZrGCozdqR}

\bibitem{winata_2021}
Winata, G.I., Madotto, A., Lin, Z., Liu, R., Yosinski, J., Fung, P.: Language
  models are few-shot multilingual learners. arXiv preprint arXiv:2109.07684
  (2021)

\bibitem{zhang2020rapidly}
Zhang, E., Gupta, N., Nogueira, R., Cho, K., Lin, J.: Rapidly deploying a
  neural search engine for the covid-19 open research dataset. In: Proceedings
  of the 1st Workshop on NLP for COVID-19 at ACL 2020 (2020)

\bibitem{metalmm}
Zhong, R., Lee, K., Zheng~Zhang, D.K.: Adapting language models for zero-shot
  learning by meta-tuning on dataset and prompt collections. arXiv preprint
  arXiv:2104.04670  (2021)

\end{thebibliography}
\bibliographystyle{splncs04}

\end{document}